\documentclass{bmvc2k}

\usepackage{graphicx}
\usepackage{comment}
\usepackage{amsmath,amssymb} 
\usepackage{color}
\usepackage{algorithm}
\usepackage{algorithmic}
\usepackage{subfigure}
\usepackage{eso-pic}
\usepackage{multirow}
\usepackage{booktabs}
\usepackage{ulem}

\title{BriNet: Towards Bridging the Intra-class and Inter-class Gaps in One-Shot Segmentation}

\addauthor{Xianghui Yang}{xianghui.yang@sydney.edu.au}{1}
\addauthor{Bairun Wang}{wangbairun@sensetime.com}{2}
\addauthor{Kaige Chen}{chenkaige@sensetime.com}{2}
\addauthor{Xinchi Zhou}{xinchi.zhou1@sydney.edu.au}{1}
\addauthor{Shuai Yi}{yishuai@sensetime.com}{2}
\addauthor{Wanli Ouyang}{wanli.ouyang@sydney.edu.au}{1}
\addauthor{Luping Zhou}{luping.zhou@sydney.edu.au}{1}

\addinstitution{
 The University of Sydney\\
 Sydney, AU
}
\addinstitution{
 SenseTime, Inc.\\
 Beijing, CN
}

\runninghead{Yang et al.}{BRINET FOR ONE-SHOT SEGMENTATION}

\def\ie{\textit{i.e}\bmvaOneDot}

\def\etal{\textit{et al}\bmvaOneDot}
\def\net{BriNet}
\begin{document}

\maketitle

\begin{abstract}
 Few-shot segmentation focuses on the generalization of models to segment unseen object instances with limited training samples. Although tremendous improvements have been achieved, existing methods are still constrained by two factors. (1) The information interaction between query and support images is not adequate, leaving intra-class gap. (2) The object categories at the training and inference stages have no overlap, leaving the inter-class gap. Thus, we propose a framework, BriNet, to bridge these gaps. First, more information interactions are encouraged between the extracted features of the query and support images, \ie, using an Information Exchange Module to emphasize the common objects. Furthermore, to precisely localize the query objects, we design a multi-path fine-grained strategy which is able to make better use of the support feature representations. Second, a new online refinement strategy is proposed to help the trained model adapt to unseen classes, achieved by switching the roles of the query and the support images at the inference stage. The effectiveness of our framework is demonstrated by experimental results, which outperforms other competitive methods and leads to a new state-of-the-art on both PASCAL VOC and MSCOCO dataset. This project can be found at \url{https://github.com/Wi-sc/BriNet}.

\end{abstract}

\section{Introduction}
\label{sec:intro}
The past decade has witnessed the fast development of deep learning in computer vision~\cite{segnet, unet, refinenet, pspnet, deeplabv1, deeplabv2, deeplabv3, deeplabv3+, Girshick_2014, Girshick_2015, He_2017, zhou2020cheaper}. Semantic segmentation is one of the fundamental tasks in computer vision which aims at predicting the pixel-wise label of images. 
Despite the success brought by deep neural networks, the training of deep segmentation models still relies on large-scale datasets, such as ImageNet~\cite{ImageNet}, PASCAL VOC~\cite{pascal} and MSCOCO~\cite{COCO}. In some cases, large-scale datasets are hard to attain due to the image collection and annotation costs.  
Moreover, the segmentation performance decreases significantly when the trained model is applied to unseen classes of objects. To solve this problem, few-shot segmentation was proposed by Shaban \etal\cite{Shaban_2017}.

Few-shot segmentation studies how to segment the target objects in a query image given a few (even only one) support images containing the objects of the same class with ground-truth segmentation masks. Typically, few-shot segmentation models take three items as input, a support image, its segmentation mask, and a query image, at both the training (offline) and the testing (online) stages. Please note that the categories at the online stage have no intersections with those at the offline stage. 

Impressive performance has been achieved as in follow-up works~\cite{canet,panet,Prototype,fwb,pgnet,local-transform}. However, we observe the two limitations.
\textbf{First}, the interaction of query and support has not been fully exploited to handle the intra-class gap that comes from the variations of objects within the same class.
Current interaction is usually unidirectional and only utilized after feature extraction, \ie, using support image information to influence the query image attention. Besides, the support-query relationship is measured via the similarity between the averaged support features of the masked regions and the local features of the query images. But the single coarse correlation is insufficient to precisely localize the objects in query images.
\textbf{Second} and more importantly, most works directly apply the trained models to segmenting unseen categories at the test stage, without considering the inter-class gap between the training and the inference object categories. 

\begin{figure}[!ht]
\begin{center}
\includegraphics[width=12cm]{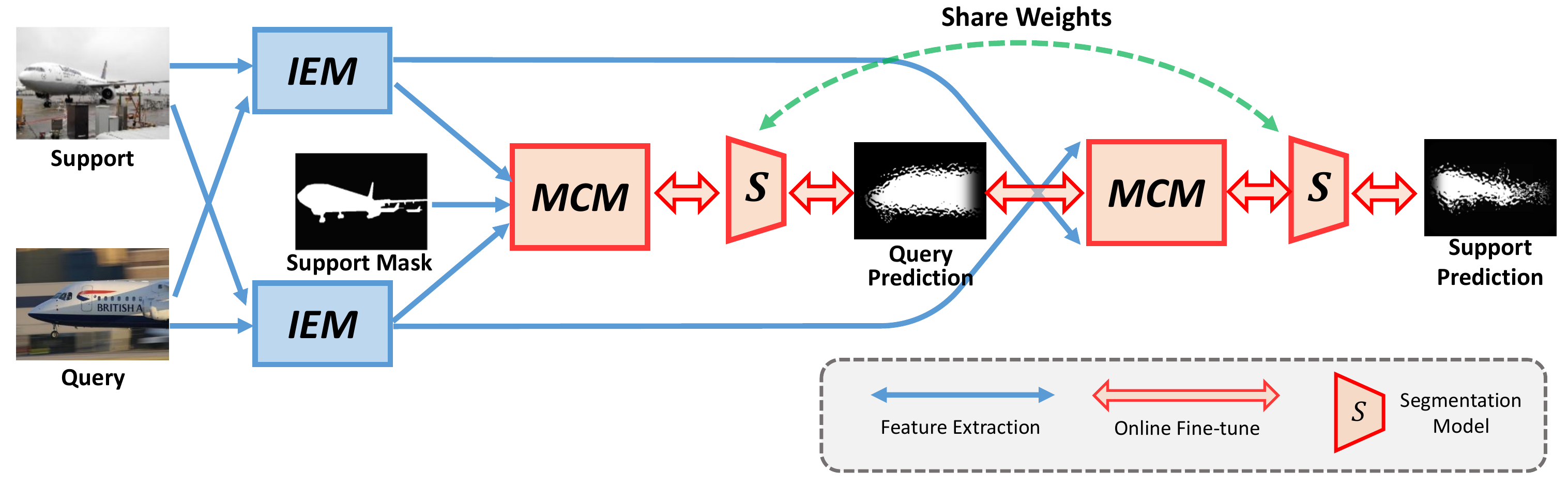}
\end{center}
\vspace{-3mm}
\caption{Overview of the proposed {\net} framework under one-shot segmentation scenario. The proposed model takes a query image and a support image with its segmentation mask as the input. The offline model consists of 2 novel modules, Information Exchange Module (IEM) and Multi-scale Correlation Module (MCM). At the online refinement stage, the roles of the query and the support images are switched, and the model is tuned to predict the segmentation mask of the support image that has the ground-truth, given the query image and the estimated query mask obtained from the initially trained model.}
\label{fig:overview}
\end{figure}

To address the above two gaps, we propose a framework named {\net}, which differs from former works in the following aspects. \textbf{First}, to narrow the intra-class gap between the support and query images, we introduce an Information Exchange Module (IEM) that learns the non-local \textit{bi-directional} transforms between support and query images, since they contain objects of the same category. The joint learning of feature representations make the deep model focus on the similar parts, \ie the target objects to segment. 
Besides, rather than globally pooling the whole object region in a support image, 
we partition the whole object into sub-regions and conduct local pooling in each region to capture more details of the object and this process is conducted in the Multi-scale Correlation Module (MCM). 
\textbf{Second}, to effectively handle the category gap between the training and inference stages, we propose an online refinement strategy to make the network adaptive and robust to unseen object classes. The roles of query and support images are exchanged to offset the lack of labels for query images and then the network is refined by minimizing the segmentation errors of the support images with ground-truth labels. This strategy provides an additional supervision signal which effectively alleviates performance drop caused by the category gap. Our strategy is versatile and able to work well with other few-shot segmentation methods for further improvements. 
Our proposed framework outperforms other competitive methods and leads to a new state-of-the-art on both PASCAL VOC and MSCOCO dataset. Fig.~\ref{fig:overview} shows an overview of our framework for one-shot segmentation.


\section{Related Work}\label{Sec:Related-work}
\noindent\textbf{Semantic segmentation.} Semantic segmentation proceeds dense classification of each pixel in an image. Recent breakthroughs mainly benefit from deep CNNs~\cite{segnet,unet,refinenet, pspnet, deeplabv1}. The Dilated Convolution~\cite{deeplabv2, deeplabv3}, which is adopted in our work, enlarges the receptive field and boosts the segmentation performance. However, the training of deep-CNN-based segmentation models relies on large-scale datasets and once trained, the models cannot be deployed to unseen categories. Few-shot Semantic Segmentation is proposed to overcome the above issues.

\noindent\textbf{Few-shot learning.} Few-shot learning focuses on generalizing models to new classes with limited training samples. Few-shot classification, as a fundamental task, attracts lots of attention, including memory methods~\cite{santoro2016meta, munkhdalai2017meta}, online refinement~\cite{finn2017modelagnostic, ravi2016optimization}, parameter prediction~\cite{bertinetto2016learning, wang2016learning2learn}, data augmentation with generative models~\cite{schwartz2018deltaencoder, Wang_2018} and metric learning~\cite{snell2017prototypical, Sung_2018, koch2015siamese}. 
Our work is most related to online refinement. Inspired but different from former refinement strategy, we design a novel pseudo supervision subtly, which bridges the inter-class gap, specifically in the few-shot segmentation task.

\noindent\textbf{Few-shot semantic segmentation.} Few-shot semantic segmentation is firstly proposed by Shaban \etal\cite{Shaban_2017}. A common paradigm employs a 2-branch architecture where the support branch generates the classification weights and the query branch produces the segmentation results, then followed by \cite{Prototype,panet,local-transform}.
Among the following works, co-FCN~\cite{conditional} and SG-One~\cite{sgone} calculate the similarity of the query features and support features to establish the relationship between support and query images. Later on, CaNet \cite{canet} introduces an iterative refinement module to improve the prediction iteratively. Zhang \etal\cite{pgnet} model the support and query feature relation with local-local correlation, instead of the global-local one, by using attention graph network. Nguyen \etal\cite{fwb} argue that there exists some unused information in test support images so that an online boosting mechanism is proposed, where support features are updated in the evaluation process. But they still ignore the information from the test query images. However, our framework utilize ignored query information to further bridge the gaps between both training and inference stages. 


\section{Task Description}


Let $\mathcal{D}_{train}=\{(\mathbf{x}_{*}^{train},\mathbf{m}_{*}^{train})\}$ be the training set and $\mathcal{D}_{test}=\{(\mathbf{x}_*^{test},\mathbf{m}_*^{test})\}$ be the test set, where $\mathbf{x}_{*}$ and $\mathbf{m}_{*}$ denote the image set and the segmentation mask set, respectively, collected from either the query images (with the subscript $q$) or the support images (with the subscript $s$). Few-shot segmentation assumes that $\mathcal{D}_{train}\cap \mathcal{D}_{test}=\emptyset$ and each pair of the query image ${\mathbf x}_q$ and its support images set \{${\mathbf x}_s^i$\} ($i=1,\cdots,K$) has at least one common object.
Given the input triplets $({\mathbf x}_q, {\mathbf x}_s^i, {\mathbf m}_s^i)$ sampled from $\mathcal{D}_{train}$, where ${\mathbf m}_s^i$ is the binary mask of ${\mathbf x}_s^i$, few-shot segmentation estimates the query mask $\hat{\mathbf m}_q$. For simplicity, in the following, we discuss our method under the scenario of $K=1$ in Section~\ref{Sec:method}.


\section{Method}\label{Sec:method}

\begin{figure}[!ht]
\begin{center}
\includegraphics[width=12cm]{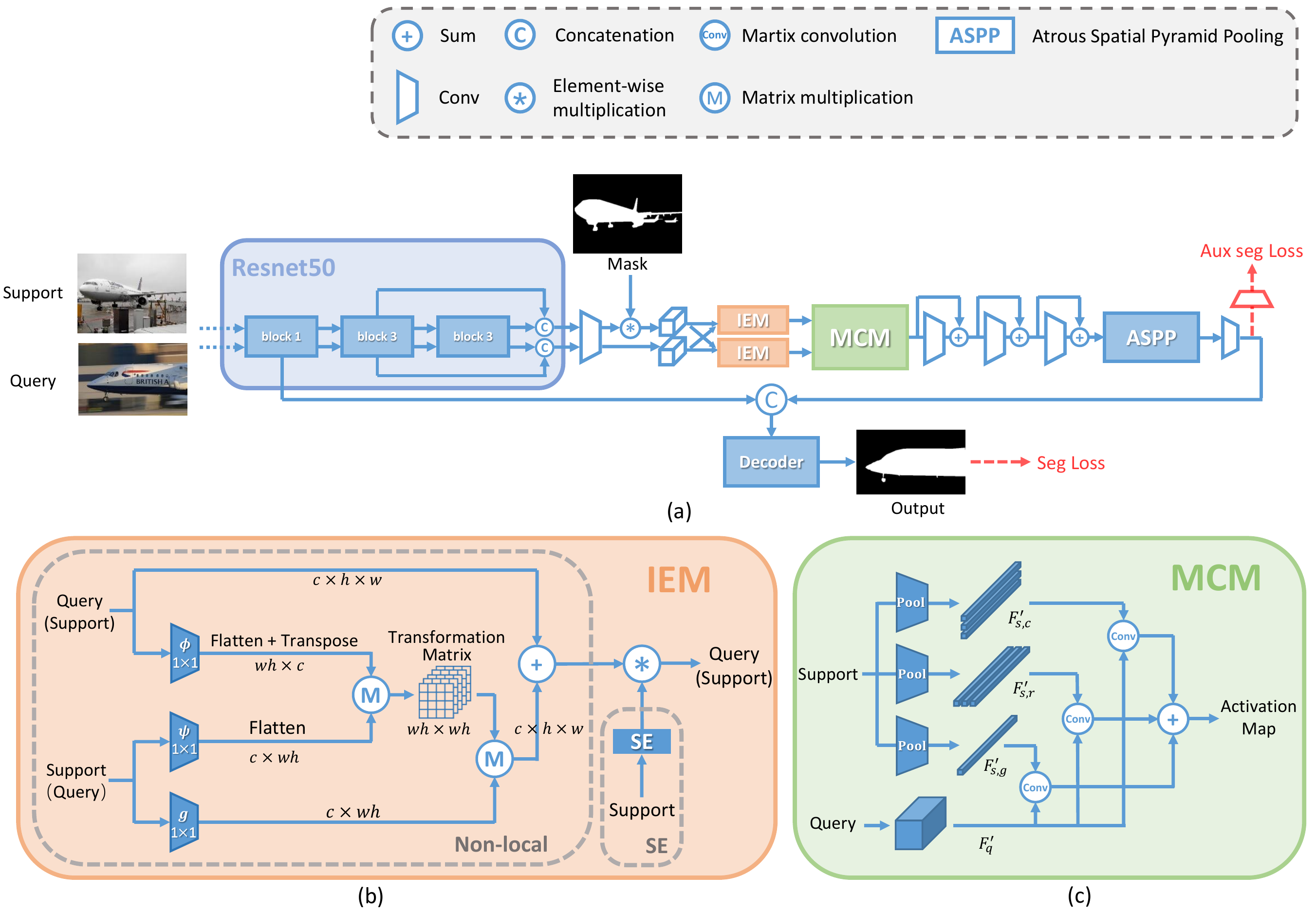}
\end{center}
\vspace{-3mm}
\caption{Network architecture of the proposed offline training model. (a) The overall architecture. (b) The detailed Information Exchange Module (IEM). (c) The detailed Multi-scale Correlation Module (MCM).}
\label{fig:offline}
\end{figure}

In this section, we present our proposed model, {\net}. It consists of an offline segmentation model and an online refinement algorithm. Specifically, for the \textbf{Offline} model, as shown in Fig.~\ref{fig:offline}, an Information Exchange Module (IEM) and a Multi-scale Correlation Module (MCM) are proposed to enhance the similarity comparison and feature fusion, respectively. For the \textbf{Online} refinement, we propose a role-switching method to adapt the trained offline model to unseen categories, which will be illustrated in Sec.~\ref{online_sec}.

\subsection{Offline Segmentation Model}

\textbf{Information Exchange Module.} Given a pair of support and query images, their features are initially extracted by a common CNN model, such as ResNet. Following that, the information exchange modules are introduced to refine the features based on our belief that the common information contained in the support and the query images should be shared and co-weighted during feature extracting. The details of the IEM are given in Fig.~\ref{fig:offline} \textcolor{red}{(b)}. 

Specifically, before the initial feature maps $F_s,~F_q \in \mathbb{R}^{c\times h\times w}$ extracted from ${\mathbf x}_s,~{\mathbf x}_q\in \mathbb{R}^{3\times H\times W}$ by CNN could be embedded by an IEM, we apply the support mask $\mathbf m_s$ on the support feature map $F_s$ to filter out unrelated background information. The resulting pure target object feature map can be written as $F_{s}^{m}=F_{s}\odot \mathbf{m_s}$,
where $\odot$ denotes element-wise multiplication and $ \mathbf m_s$ is down-sampled from ${W\times H}$ into the identical size ${w\times h}$ with $F_s$.

Our IEM takes $F_{s}^{m}$ and $F_q$ as inputs. It consists of Non-local block $\mathcal{T}$ \cite{nonlocal} and Squeeze-and-Excitation (SE) block $\mathcal{C}$ \cite{se}, where Non-local block is used for information exchange and SE block aims to boosting channel information of target object. The Non-local block was also adopted by \cite{Co-Att-Co-Excitation} for few-shot object detection but without applying foreground mask. IEM outputs the refined feature maps $F_q^{'}$ and $F_s^{'}$, formulated as Eq.~\ref{eq2},
\begin{align}
\label{eq2}
F_q^{'}=\mathcal{T}(F_q, F_{s}^{m})\odot \mathcal{C}(F_{s}^{m}), \hspace{4mm}
F_s^{'}=\mathcal{T}(F_{s}^{m}, F_q)\odot \mathcal{C}(F_{s}^{m}).
\end{align}
Non-local block is originally proposed to capture long-range dependencies focusing on the regions of a single image input, where the non-local transformation is modeled by relations among local features. In contrast, in our approach, the Non-local block under support-query input setting is for inter-input transformation, denoted as Transformation Matrix in Fig.~\ref{fig:offline} \textcolor{red}{(b)}, where all local features of one input (say, the support) are transformed into one local feature of the other input (say, the query). The support-query transformation is applied as Eq.~\ref{eq4} and Eq.~\ref{eq5},
\begin{equation}
\label{eq4}
\mathcal{T}(F_q, F_s^{m})_{i}=\sum_{j}^{wh}{\phi(F_{q})_i}^{T}\psi( F_s^{m})_{j}\cdot g(F_{s}^{m})_j+F_{q,i}
\end{equation}
\begin{equation}
\label{eq5}
\mathcal{T}(F_s^{m}, F_q)_{i}=\sum_{j}^{wh}{\phi(F_{s}^{m})_i}^{T} \psi(F_{q})_{j}\cdot g(F_{q})_j+F_{s,i}^{m}
\end{equation}
where $\phi,\psi$ and $g$ are $1\times 1$ convolution kernels and $i,j$ are the index of pixel. The SE block generates channel attention by Global Average/Max Pooling, followed by two sequential MLP layers.

\noindent\textbf{Multi-scale Correlation Module.} In contrast to previous coarse global mask pooling, our proposed method conduct region pooling in a more fine-grained manner, which achieves the balance between computation overhead and feature details. Fig.~\ref{fig:offline} \textcolor{red}{(c)} shows the details of MCM. 

Specifically, given IEM ouputs $F_s^{'},F_q^{'}\in \mathbb{R}^{c\times h\times w}$, apart from global average pooling, we apply 2 slide average pooling windows on feature map $F_{s}^{'}\in \mathbb{R}^{c\times w\times h}$ with size $s\times h, w\times s$ and strides $s, s$ respectively, where $s$ is the stride size. As a result, more fine-grained feature representations $F_{s,c}^{'}\in \mathbb{R}^{c\times \frac{w}{s}\times 1}, F_{s,r}^{'}\in \mathbb{R}^{c\times 1\times \frac{h}{s}}, F_{s,g}^{'}\in \mathbb{R}^{c\times 1\times 1}$ are obtained. After convoluting the query feature map $F_{q}^{'}\in \mathbb{R}^{c\times w\times h}$ with these 3 kernels respectively, the three activation maps are summed as Eq.~\ref{eq6},
\begin{equation}
\label{eq6}
F_{s-q}=F_{s,c}^{'}*F_{q}^{'}+F_{s,r}^{'}*F_{q}^{'}+F_{s,g}^{'}*F_{q}^{'}
\end{equation}
where $*$ denotes convolution operation and $F_{s-q}$ is the output of MCM.

\textbf{Loss function}. To boost performance, in addition to the final cross-entropy segmentation loss $\mathcal{L}_{seg}$, we introduce another auxiliary segmentation branch into our proposed architecture before the Decoder to shorten gradient propagation, as shown in Fig.~\ref{fig:offline}. The auxiliary cross-entropy segmentation loss $\mathcal{L}_{aux-seg}$ is minimized together with the final segmentation loss $\mathcal{L}_{seg}$. Thus the overall loss function is
\begin{equation}\nonumber
\mathcal{L}=\mathcal{L}_{seg}+\mathcal{L}_{aux-seg}
\end{equation}

\subsection{Online Refinement}
\label{online_sec}


In order to make our model adaptive to the agnostic objects in the test stage, we propose to conduct online refinement. Our proposed online refinement algorithm takes advantage of the support-query pair available at the test stage and switches their roles to extract complementary information for the refinement iteratively.

According to the definition of few-shot segmentation, at test stage, the information about the agnostic categories in the support image $x_s^{test}$, as well as its corresponding mask $m_s^{test}$, and query image $x_q^{test}$ are available, but the query mask $m_q^{test}$ and other support-query pairs from $\mathcal{D}_{test}$ are unknown. Inspired by self-supervision, the core of our online refinement is to regard the query offline prediction $\hat{m_q^{test}}$ as the pseudo (ground-truth) mask to assist the support image segmentation in return. This refinement could be conducted for a few rounds by switching the roles of the query and the support images iteratively.


Specifically,  given a query image $x_q^{test}$, a support image $x_s^{test}$ and a support ground-truth mask $m_s^{test}$, we do the online refinement in three steps. First, we feed $x_q^{test}$, $x_s^{test}$ and $m_s^{test}$ into the model obtained at the offline training to estimate the query mask $\hat{m_q^{test}}$. Second, $\hat{m_q^{test}}$ is treated as the pseudo ground-truth label for $x_q^{test}$, which constitutes the next input with $x_q^{test}$, $x_s^{test}$ to predict the support mask $\hat{m_s^{test}}$. Third, the segmentation model is constantly refined by minimizing the cross-entropy loss between the predicted support mask $\hat{m_s^{test}}$ and the ground-truth support mask $m_s^{test}$. These 3 steps are repeated until the mean IoU between $m_s^{test}$ and $\hat{m_s^{test}}$ is higher than a threshold $t_i=t_0*\frac{N-1}{N+i}$ in the $i$-th step or the maximum iteration time $N$ has been reached. This algorithm is formulated as alternatively updating $\hat{m_s}$ and $\hat{m_q}$ according to Eq.~\ref{mcm},
\begin{align}
\label{mcm}
\hat{m_s}=\mathcal{F}(x_s, x_q, \hat{m_q}), \hspace{4mm}
\hat{m_q}=\mathcal{F}(x_q, x_s, m_s).
\end{align}
Here $\mathcal{F}$ indicates the embedding function of the model either trained offline or refined online in the last iteration. 

\section{Experiments}\label{Sec:exp}

\subsection{Datasets and Evaluation metric} 
We evaluate the performance of our model on two benchmark datasets commonly used for few-shot segmentation.

\noindent\textbf{PASCAL-$\mathbf{5^i}$.} This dataset was composed based on PASCAL VOC 2012~\cite{pascal} and the extended SDS datasets~\cite{sds}. Following the work in~\cite{Shaban_2017} and the conventional evaluation on this dataset, we adopt 4-fold cross validation that divides the 20 classes of PASCAL into four folds, three of which are used for training and the rest one for test. It is noted that the selection of support and query image pairs could influence the performance. Following~\cite{Shaban_2017}, we randomly sample the support and query image pairs 1000 times from the test set for evaluation. 

\noindent\textbf{COCO-$\mathbf{{20}^i}$.} Up to now, one-shot segmentation mainly takes PASCAL-$5^i$ for evaluation. Only Zhang \etal~\cite{canet} and Nguyen \etal~\cite{fwb} tested their methods on MSCOCO, and their dataset settings are even not the same. Zhang \etal~\cite{canet} divide 80 classes into three parts, of which 40 classes are used for training, 20 classes for evaluation and the remaining 20 classes for test. Nguyen \etal~\cite{fwb} processed the dataset as the PASCAL-$5^i$, where the 80 classes are divided into 4 folds and each fold contains 20 classes, named COCO-${20}^i$. We follow~\cite{fwb} to evaluate our model on COCO-${20}^i$. 

\noindent\textbf{Evaluation metrics.} To be consistent with the literature for comparison, class related foreground Intersection-over-Union (F-IoU) is adopted in this paper, which is computed as follows. First, the foreground intersection and union pixel numbers are summed according to classes; Second, the foreground Intersection-over-Union ratio is computed for each class; Third, the average IoU over all classes (mean IoU) are reported as the evaluation metric to reveal the overall performance.

\subsection{Implementation details}

We adopt ResNet50~\cite{resnet} modified from Deeplab V3~\cite{deeplabv3} as our model backbone.
In view of the task characteristic, we abandon ResBlock-4 and the later layers of ResNet50, which is consistent with the existing works in~\cite{canet,pgnet}. The parameters of ResNet are initialized from the model pre-trained by ImageNet~\cite{ImageNet} and fixed during training.

 Our model is trained using SGD for 400 epochs on Nvidia Titan V GPUs. We set the base learning rate to 2e-2 and reduce it to 2e-3 after 150 epochs. Momentum and weight decay of SGD are set to 0.9 and 5e-4, respectively. The input images have the size of $353\times 353$. For data augmentation, we follow CaNet~\cite{canet} to adopt random mirror, random rotation, random resize and random crop for both datasets. For online fine-tune, the iteration number is $20$.

\subsection{Results}

\begin{table}[ht]
\begin{center}
{\small
\begin{tabular}{c|c|c|c|c|c|c}
\hline
\bf Methods & \bf Backbone & {\bf fold-0} & {\bf fold-1} & {\bf fold-2} & {\bf fold-3} & {\bf Mean}\\
\hline
FW\&B~\cite{fwb} & VGG16 & 18.35 & 16.72 & 19.59 & 25.43 & 20.2\\
\hline
FW\&B~\cite{fwb} & ResNet101 & 16.98 & 17.98 & 20.96 & 28.85 & 21.19 \\
\hline
PGNet* & \multirow{2}*{ResNet50} & 32.24 & 30.51 & 31.61 & 29.73 & 31.02 \\
CaNet* & ~ & \bf 34.25 & 34.44 & 30.87 &\bf 31.21 & 32.69 \\
\hline
 Ours Offline & \multirow{2}*{ResNet50}& 31.40 &  36.01 & 36.78 & 29.86 & 33.51 \\
 Ours Offline + Online & ~ & 32.88 & \bf 36.20 & \bf 37.44 & 30.93 & \bf 34.36 \\
\hline
\end{tabular}
}
\end{center}
\vspace{-2mm}
\caption{Comparison with the state-of-the-art 1-shot segmentation performance on COCO-$20^i$. The symbol * indicates the model is re-run by ourselves.}
\label{table:coco-f}

\begin{center}
{\small
\begin{tabular}{c|c|c|c|c|c|c}
\hline
\bf Methods & \bf Backbone & {\bf fold-0} & {\bf fold-1} & {\bf fold-2} & {\bf fold-3} & {\bf Mean}\\
\hline
OSLSM~\cite{Shaban_2017} & \multirow{6}*{VGG-16} & 33.6 & 55.3 & 40.9 & 33.5 & 40.8\\
co-FCN~\cite{conditional} & ~ & 36.7 & 50.6 & 44.9 & 32.4 & 41.1\\
PL+SEG+PT~\cite{Prototype} & ~ & - & - & - & - & 42.7\\
AMP~\cite{AMP} & ~ & 41.9 & 50.2 & 46.7 & 34.4 & 43.4\\
SG-One~\cite{sgone} & ~ & 40.2 & 58.4 & 48.4 & 38.4 & 46.3\\
PANet~\cite{panet} & ~ & 42.3 & 58.0 & 51.1 & 41.2 & 48.1 \\
\hline
CaNet~\cite{canet} & \multirow{4}*{ResNet50} & 52.5 & 65.9 & 51.3 & 51.9 & 55.4\\
PGNet~\cite{pgnet} & ~ & 56.0 & 66.9 & 50.6 & 50.4 & 56.0 \\
CaNet* & ~ & 51.11 & 66.09 & 50.06 & 52.57 & 54.96\\
PGNet* & ~ & 53.63 & 65.70 & 48.54 & 49.28 & 54.29 \\
\hline
FW\&B~\cite{fwb} & ResNet-101 &  51.3 & 64.5 & 56.7 & 52.2 & 56.2 \\
\hline
Ours Offline & \multirow{2}*{ResNet-50} &\bf 56.85 & \bf 67.52 & 48.89 &\bf 53.23 & 56.62 \\
Ours Offline + Online & ~  & 56.54 & 67.20 &\bf 51.56 & 53.02 &\bf 57.08\\
\hline
\end{tabular}
}
\end{center}
\vspace{-2mm}
\caption{Comparison with the state-of-the-art 1-shot segmentation performance on PASCAL-$5^i$. The symbol * indicates the model is re-run by ourselves.}
\label{table:pascal-f}
\end{table}


We evaluate the performance of our proposed model with/without online refinement, and compare it with multiple state-of-the-art methods for few-shot segmentation. The results are reported in Tab.~\ref{table:coco-f} and Tab.~\ref{table:pascal-f}. It is worth mentioning that, for the two closely related methods CaNet~\cite{canet} and PGNet~\cite{pgnet}, in addition to directly quoting the results from the original papers (for PASCAL-$5^i$), we also re-run the models and report the results, indicated as CaNet* and PGNet* in the tables. For CaNet* and PGNet*,  the same support-query pairs are used for test as in our model. In this way, a strict comparison that further removes the difference in randomly sampling test pairs is conducted. 

The results on COCO-$20^i$ are given in Tab.~\ref{table:coco-f}. COCO-$20^i$ is a very challenging dataset. As can be seen, on this task, even our offline model only has outperformed the four competitors in terms of the mean IoU. With the proposed online refinement, the performance of our model could be further boosted, making its advantage over the other methods in comparison more salient.
The results on PASCAL-$5^i$ are given in Tab.~\ref{table:pascal-f}. This is a relatively easy task and all methods in comparison have better performance than what they do on COCO-$20^i$. The performance of our offline model is comparable to that of the second best method FW\&B which builds on a more powerful backbone network ResNet-101.  Compared with CaNet* and PGNet* that use the same test pairs as ours, our offline model wins both of them with a large margin. Again, our online refinement could further improve our performance on this dataset consistently. Moreover, cross-referencing the results in Tab.~\ref{table:coco-f} and Tab.~\ref{table:pascal-f}, it seems that our online refinement contributes more to the performance improvement when the segmentation task is hard. 

\begin{figure}[ht]
\begin{center}
\includegraphics[width=12cm]{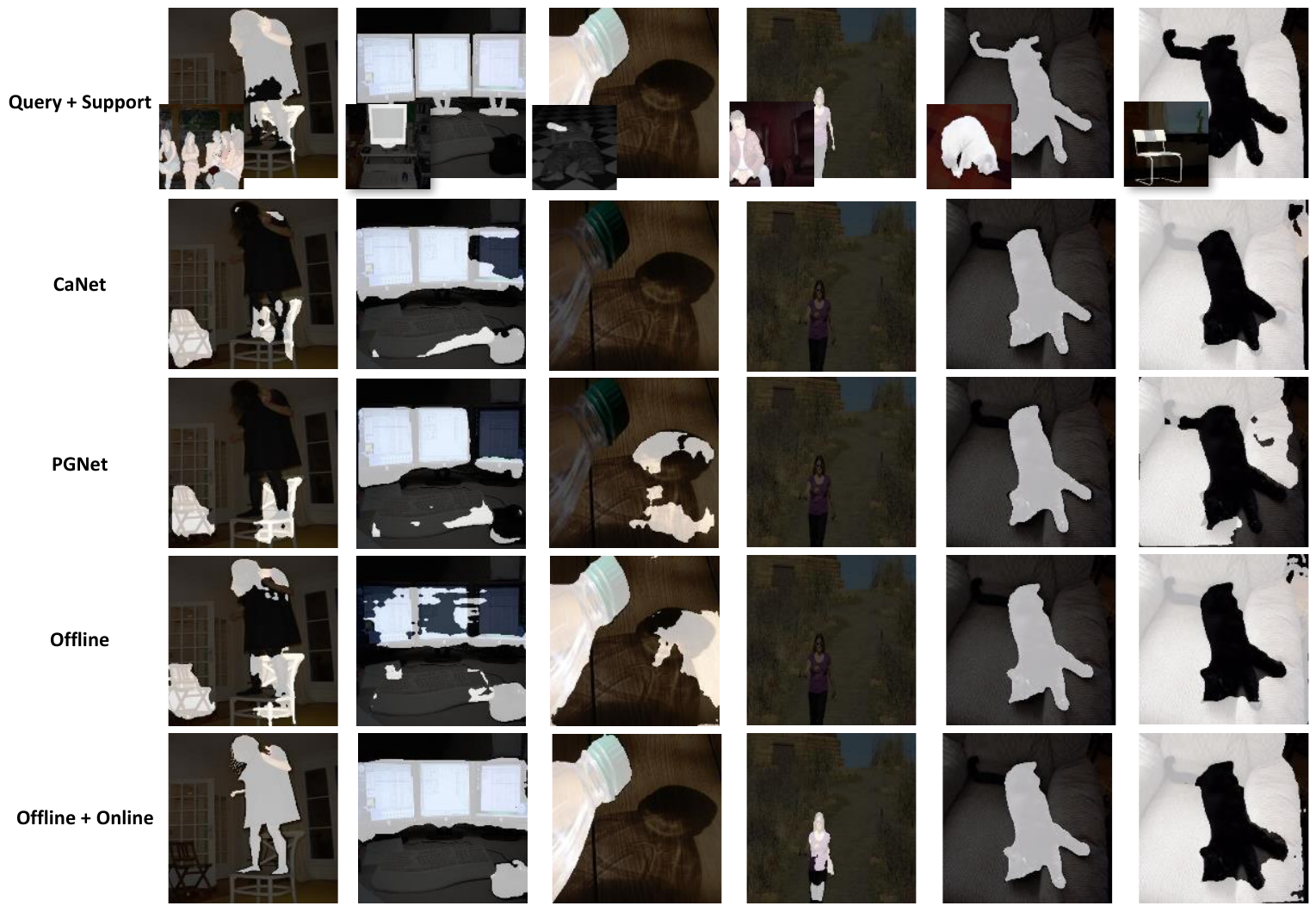}
\end{center}
\vspace{-3mm}
\caption{Six visual examples (corresponding to six columns) from PASCAL-$5^i$ under 1-shot segmentation. 1st row: the query images and support images (in the small window) with ground-truth segmentation. 2nd-5th rows: the query images and segmentation predicted by CaNet*, PGNet*, Offline and Offline + Online models, respectively. Our BriNet outperforms previous best frameworks and our online refinement algorithm improves the segmentation performance of offline model significantly. Best viewed in color.}
\label{fig:visualization}
\end{figure}

Fig.~\ref{fig:visualization} shows six visual examples of segmentation results from our proposed BriNet and previous best models, CaNet and PGNet. Given the same query image, all of CaNet, PGNet and our BriNet are able to segment different classes with different support examples as guidance (the two rightmost columns in Fig.~\ref{fig:visualization}). However, our BriNet can generate more accurate and complete segmentation results compared with CaNet and PGNet, even when both of them totally fail (3rd column and 4th column). Our online refinement improves the model adaptation to agnostic object segmentation significantly (the last two rows in Fig.~\ref{fig:visualization}).



\subsection{Ablation Study}\label{Sec:ablation}
To single out the contribution of each component proposed in our model, we conduct an ablation study in order to answer two questions: (i) How do IEM and MCM contribute to the performance of the offline model? (ii) Could our online refinement, as a general method, help other few-shot segmentation models improve the performance? 4-fold validation is used and the mean IoU values are reported.


\begin{table}
\begin{center}
    \begin{minipage}[t]{0.48\textwidth}
    \begin{center}
    {\small
    \begin{tabular}{c|c|c}
    \hline
    \bf Model &{\bf COCO}& {\bf PASCAL}\\
    \hline
     BriNet w/o IEM  & 28.08 & 53.25 \\
    BriNet w/o MCM  & 30.64 & 53.49 \\
    \hline
    BriNet & 33.51 & 56.62 \\
    \hline
    \end{tabular}
    }
    \end{center}
    \vspace{-2mm}
    \caption{Ablation study about IEM and MCM on COCO and PASCAL.}
    \label{table:offline-ablation}
    \end{minipage}\hfill\begin{minipage}[t]{0.48\textwidth}
    \begin{center}
    {\small
    \begin{tabular}{c|c|c}
    \hline
    \bf Model & {\bf COCO} & {\bf PASCAL}\\
    \hline
    CaNet*  & 32.69 & 54.96 \\
    CaNet* + Online & 32.84 & 54.92 \\
    \hline
    PGNet*  & 31.02 & 54.29 \\
    PGNet* + Online & 31.89 & 54.71 \\
    \hline
    \end{tabular}
    }
    \end{center}
    \vspace{-2mm}
    \caption{Ablation study about our online refinement.}
    \label{table:online-ablation}
    \end{minipage}
\end{center}
\end{table}

\textbf{IEM and MCM.} To answer the first question, we compare our offline model with its variants that remove IEM and MCM, respectively. The results are given in Tab.~\ref{table:offline-ablation}. As seen, without either IEM or MCM, the performance of the offline model will significantly decrease on both PASCAL-$5^i$ and COCO-$20^i$, showing the necessity of employing these two modules, as we argued before.


\textbf{Online refinement.} To answer the second question, we apply our online refinement to CaNet* and PGNet*, and the results are in Tab.~\ref{table:online-ablation}. Significant improvement could be observed on the hard classes in COCO-${20^i}$ for both models. On PASCAL-${5^i}$, although little effect is observed on CaNet*, our online refinement could help PGNet* to improve further. This experiment demonstrates the value of our online refinement method as a general strategy to improve few-shot segmentation. 

\section{Conclusions}

In this paper we proposed {\net}, a novel framework for segmentation network with few-shot learning. Our model contributes the state-of-the-arts as follows. \textbf{First}, we introduce an information exchange module to boost the feature representations of the support and query images both. Besides, we represent the masked objects in the support image in a relatively more fine-grained way to better localize the objects in the query image. \textbf{Second}, we propose a new online refinement strategy to adapt the trained model to unseen test objects. Specifically, we tactically switch the roles of the query and the support images at the test stage and refine our model by minimizing the segmentation errors of the support images. In this way, we fully exploit the additional information in both the test query image and its supporters, which has not been well handled in the existing methods. The effectiveness of our model has been demonstrated in our experiment, which outperforms the state-of-the-arts methods by a margin.

\newpage
\bibliography{egbib}

\begin{thebibliography}{41}
\providecommand{\natexlab}[1]{#1}
\providecommand{\url}[1]{\texttt{#1}}
\expandafter\ifx\csname urlstyle\endcsname\relax
  \providecommand{\doi}[1]{doi: #1}\else
  \providecommand{\doi}{doi: \begingroup \urlstyle{rm}\Url}\fi

\bibitem[Badrinarayanan et~al.(2017)Badrinarayanan, Kendall, and
  Cipolla]{segnet}
Vijay Badrinarayanan, Alex Kendall, and Roberto Cipolla.
\newblock Segnet: A deep convolutional encoder-decoder architecture for image
  segmentation.
\newblock \emph{IEEE Transactions on Pattern Analysis and Machine
  Intelligence}, 39\penalty0 (12):\penalty0 2481–2495, Dec 2017.
\newblock ISSN 1939-3539.
\newblock \doi{10.1109/tpami.2016.2644615}.
\newblock URL \url{http://dx.doi.org/10.1109/TPAMI.2016.2644615}.

\bibitem[Bertinetto et~al.(2016)Bertinetto, Henriques, Valmadre, Torr, and
  Vedaldi]{bertinetto2016learning}
Luca Bertinetto, Jo{\~a}o~F Henriques, Jack Valmadre, Philip Torr, and Andrea
  Vedaldi.
\newblock Learning feed-forward one-shot learners.
\newblock In \emph{Advances in neural information processing systems}, pages
  523--531, 2016.

\bibitem[Chen et~al.(2014)Chen, Papandreou, Kokkinos, Murphy, and
  Yuille]{deeplabv1}
Liang-Chieh Chen, George Papandreou, Iasonas Kokkinos, Kevin Murphy, and
  Alan~L. Yuille.
\newblock Semantic image segmentation with deep convolutional nets and fully
  connected crfs, 2014.

\bibitem[Chen et~al.(2017)Chen, Papandreou, Schroff, and Adam]{deeplabv3}
Liang-Chieh Chen, George Papandreou, Florian Schroff, and Hartwig Adam.
\newblock Rethinking atrous convolution for semantic image segmentation, 2017.

\bibitem[Chen et~al.(2018{\natexlab{a}})Chen, Papandreou, Kokkinos, Murphy, and
  Yuille]{deeplabv2}
Liang-Chieh Chen, George Papandreou, Iasonas Kokkinos, Kevin Murphy, and
  Alan~L. Yuille.
\newblock Deeplab: Semantic image segmentation with deep convolutional nets,
  atrous convolution, and fully connected crfs.
\newblock \emph{IEEE Transactions on Pattern Analysis and Machine
  Intelligence}, 40\penalty0 (4):\penalty0 834–848, Apr 2018{\natexlab{a}}.
\newblock ISSN 2160-9292.
\newblock \doi{10.1109/tpami.2017.2699184}.
\newblock URL \url{http://dx.doi.org/10.1109/TPAMI.2017.2699184}.

\bibitem[Chen et~al.(2018{\natexlab{b}})Chen, Zhu, Papandreou, Schroff, and
  Adam]{deeplabv3+}
Liang-Chieh Chen, Yukun Zhu, George Papandreou, Florian Schroff, and Hartwig
  Adam.
\newblock Encoder-decoder with atrous separable convolution for semantic image
  segmentation.
\newblock \emph{Lecture Notes in Computer Science}, page 833–851,
  2018{\natexlab{b}}.
\newblock ISSN 1611-3349.
\newblock \doi{10.1007/978-3-030-01234-2_49}.
\newblock URL \url{http://dx.doi.org/10.1007/978-3-030-01234-2_49}.

\bibitem[Dong and Xing(2018)]{Prototype}
Nanqing Dong and Eric~P. Xing.
\newblock Few-shot semantic segmentation with prototype learning.
\newblock In \emph{BMVC}, 2018.

\bibitem[Everingham et~al.(2010)Everingham, Van~Gool, Williams, Winn, and
  Zisserman]{pascal}
M.~Everingham, L.~Van~Gool, C.~K.~I. Williams, J.~Winn, and A.~Zisserman.
\newblock The pascal visual object classes (voc) challenge.
\newblock \emph{International Journal of Computer Vision}, 88\penalty0
  (2):\penalty0 303--338, June 2010.

\bibitem[Finn et~al.(2017)Finn, Abbeel, and Levine]{finn2017modelagnostic}
Chelsea Finn, Pieter Abbeel, and Sergey Levine.
\newblock Model-agnostic meta-learning for fast adaptation of deep networks,
  2017.

\bibitem[Girshick(2015)]{Girshick_2015}
Ross Girshick.
\newblock Fast r-cnn.
\newblock \emph{2015 IEEE International Conference on Computer Vision (ICCV)},
  Dec 2015.
\newblock \doi{10.1109/iccv.2015.169}.
\newblock URL \url{http://dx.doi.org/10.1109/ICCV.2015.169}.

\bibitem[Girshick et~al.(2014)Girshick, Donahue, Darrell, and
  Malik]{Girshick_2014}
Ross Girshick, Jeff Donahue, Trevor Darrell, and Jitendra Malik.
\newblock Rich feature hierarchies for accurate object detection and semantic
  segmentation.
\newblock \emph{2014 IEEE Conference on Computer Vision and Pattern
  Recognition}, Jun 2014.
\newblock \doi{10.1109/cvpr.2014.81}.
\newblock URL \url{http://dx.doi.org/10.1109/CVPR.2014.81}.

\bibitem[{Hariharan} et~al.(2011){Hariharan}, {Arbeláez}, {Bourdev}, {Maji},
  and {Malik}]{sds}
B.~{Hariharan}, P.~{Arbeláez}, L.~{Bourdev}, S.~{Maji}, and J.~{Malik}.
\newblock Semantic contours from inverse detectors.
\newblock In \emph{2011 International Conference on Computer Vision}, pages
  991--998, 2011.

\bibitem[He et~al.(2016)He, Zhang, Ren, and Sun]{resnet}
Kaiming He, Xiangyu Zhang, Shaoqing Ren, and Jian Sun.
\newblock Deep residual learning for image recognition.
\newblock \emph{2016 IEEE Conference on Computer Vision and Pattern Recognition
  (CVPR)}, Jun 2016.
\newblock \doi{10.1109/cvpr.2016.90}.
\newblock URL \url{http://dx.doi.org/10.1109/cvpr.2016.90}.

\bibitem[He et~al.(2017)He, Gkioxari, Dollar, and Girshick]{He_2017}
Kaiming He, Georgia Gkioxari, Piotr Dollar, and Ross Girshick.
\newblock Mask r-cnn.
\newblock \emph{2017 IEEE International Conference on Computer Vision (ICCV)},
  Oct 2017.
\newblock \doi{10.1109/iccv.2017.322}.
\newblock URL \url{http://dx.doi.org/10.1109/ICCV.2017.322}.

\bibitem[Hsieh et~al.(2019)Hsieh, Lo, Chen, and Liu]{Co-Att-Co-Excitation}
Ting-I Hsieh, Yi-Chen Lo, Hwann-Tzong Chen, and Tyng-Luh Liu.
\newblock One-shot object detection with co-attention and co-excitation, 2019.

\bibitem[Hu et~al.(2017)Hu, Shen, and Sun]{se}
Jie Hu, Li~Shen, and Gang Sun.
\newblock Squeeze-and-excitation networks.
\newblock \emph{CoRR}, abs/1709.01507, 2017.
\newblock URL \url{http://arxiv.org/abs/1709.01507}.

\bibitem[Koch et~al.(2015)Koch, Zemel, and Salakhutdinov]{koch2015siamese}
Gregory Koch, Richard Zemel, and Ruslan Salakhutdinov.
\newblock Siamese neural networks for one-shot image recognition.
\newblock In \emph{ICML deep learning workshop}, volume~2. Lille, 2015.

\bibitem[Lin et~al.(2017)Lin, Milan, Shen, and Reid]{refinenet}
Guosheng Lin, Anton Milan, Chunhua Shen, and Ian Reid.
\newblock Refinenet: Multi-path refinement networks for high-resolution
  semantic segmentation.
\newblock \emph{2017 IEEE Conference on Computer Vision and Pattern Recognition
  (CVPR)}, Jul 2017.
\newblock \doi{10.1109/cvpr.2017.549}.
\newblock URL \url{http://dx.doi.org/10.1109/CVPR.2017.549}.

\bibitem[Lin et~al.(2014)Lin, Maire, Belongie, Hays, Perona, Ramanan, Dollár,
  and Zitnick]{COCO}
Tsung-Yi Lin, Michael Maire, Serge Belongie, James Hays, Pietro Perona, Deva
  Ramanan, Piotr Dollár, and C.~Lawrence Zitnick.
\newblock Microsoft coco: Common objects in context.
\newblock \emph{Lecture Notes in Computer Science}, page 740–755, 2014.
\newblock ISSN 1611-3349.
\newblock \doi{10.1007/978-3-319-10602-1_48}.
\newblock URL \url{http://dx.doi.org/10.1007/978-3-319-10602-1_48}.

\bibitem[Munkhdalai and Yu(2017)]{munkhdalai2017meta}
Tsendsuren Munkhdalai and Hong Yu.
\newblock Meta networks, 2017.

\bibitem[Nguyen and Todorovic(2019)]{fwb}
Khoi Nguyen and Sinisa Todorovic.
\newblock Feature weighting and boosting for few-shot segmentation.
\newblock \emph{ArXiv}, abs/1909.13140, 2019.

\bibitem[Rakelly et~al.(2018)Rakelly, Shelhamer, Darrell, Efros, and
  Levine]{conditional}
Kate Rakelly, Evan Shelhamer, Trevor Darrell, Alexei~A. Efros, and Sergey
  Levine.
\newblock Few-shot segmentation propagation with guided networks, 2018.

\bibitem[Ravi and Larochelle(2016)]{ravi2016optimization}
Sachin Ravi and Hugo Larochelle.
\newblock Optimization as a model for few-shot learning.
\newblock 2016.

\bibitem[Ronneberger et~al.(2015)Ronneberger, Fischer, and Brox]{unet}
Olaf Ronneberger, Philipp Fischer, and Thomas Brox.
\newblock U-net: Convolutional networks for biomedical image segmentation.
\newblock \emph{Medical Image Computing and Computer-Assisted Intervention –
  MICCAI 2015}, page 234–241, 2015.
\newblock ISSN 1611-3349.
\newblock \doi{10.1007/978-3-319-24574-4_28}.
\newblock URL \url{http://dx.doi.org/10.1007/978-3-319-24574-4_28}.

\bibitem[Russakovsky et~al.(2015)Russakovsky, Deng, Su, Krause, Satheesh, Ma,
  Huang, Karpathy, Khosla, Bernstein, and et~al.]{ImageNet}
Olga Russakovsky, Jia Deng, Hao Su, Jonathan Krause, Sanjeev Satheesh, Sean Ma,
  Zhiheng Huang, Andrej Karpathy, Aditya Khosla, Michael Bernstein, and et~al.
\newblock Imagenet large scale visual recognition challenge.
\newblock \emph{International Journal of Computer Vision}, 115\penalty0
  (3):\penalty0 211–252, Apr 2015.
\newblock ISSN 1573-1405.
\newblock \doi{10.1007/s11263-015-0816-y}.
\newblock URL \url{http://dx.doi.org/10.1007/s11263-015-0816-y}.

\bibitem[Santoro et~al.(2016)Santoro, Bartunov, Botvinick, Wierstra, and
  Lillicrap]{santoro2016meta}
Adam Santoro, Sergey Bartunov, Matthew Botvinick, Daan Wierstra, and Timothy
  Lillicrap.
\newblock Meta-learning with memory-augmented neural networks.
\newblock In \emph{International conference on machine learning}, pages
  1842--1850, 2016.

\bibitem[Schwartz et~al.(2018)Schwartz, Karlinsky, Shtok, Harary, Marder,
  Feris, Kumar, Giryes, and Bronstein]{schwartz2018deltaencoder}
Eli Schwartz, Leonid Karlinsky, Joseph Shtok, Sivan Harary, Mattias Marder,
  Rogerio Feris, Abhishek Kumar, Raja Giryes, and Alex~M. Bronstein.
\newblock Delta-encoder: an effective sample synthesis method for few-shot
  object recognition, 2018.

\bibitem[Shaban et~al.(2017)Shaban, Bansal, Liu, Essa, and Boots]{Shaban_2017}
Amirreza Shaban, Shray Bansal, Zhen Liu, Irfan Essa, and Byron Boots.
\newblock One-shot learning for semantic segmentation.
\newblock \emph{Procedings of the British Machine Vision Conference 2017},
  2017.
\newblock \doi{10.5244/c.31.167}.
\newblock URL \url{http://dx.doi.org/10.5244/c.31.167}.

\bibitem[Siam et~al.(2019)Siam, Oreshkin, and Jagersand]{AMP}
Mennatullah Siam, Boris Oreshkin, and Martin Jagersand.
\newblock Adaptive masked proxies for few-shot segmentation, 2019.

\bibitem[Snell et~al.(2017)Snell, Swersky, and Zemel]{snell2017prototypical}
Jake Snell, Kevin Swersky, and Richard~S. Zemel.
\newblock Prototypical networks for few-shot learning, 2017.

\bibitem[Sung et~al.(2018)Sung, Yang, Zhang, Xiang, Torr, and
  Hospedales]{Sung_2018}
Flood Sung, Yongxin Yang, Li~Zhang, Tao Xiang, Philip~H.S. Torr, and Timothy~M.
  Hospedales.
\newblock Learning to compare: Relation network for few-shot learning.
\newblock \emph{2018 IEEE/CVF Conference on Computer Vision and Pattern
  Recognition}, Jun 2018.
\newblock \doi{10.1109/cvpr.2018.00131}.
\newblock URL \url{http://dx.doi.org/10.1109/CVPR.2018.00131}.

\bibitem[Wang et~al.(2019)Wang, Liew, Zou, Zhou, and Feng]{panet}
Kaixin Wang, Jun~Hao Liew, Yingtian Zou, Daquan Zhou, and Jiashi Feng.
\newblock Panet: Few-shot image semantic segmentation with prototype alignment,
  2019.

\bibitem[Wang et~al.(2018{\natexlab{a}})Wang, Girshick, Gupta, and
  He]{nonlocal}
Xiaolong Wang, Ross Girshick, Abhinav Gupta, and Kaiming He.
\newblock Non-local neural networks.
\newblock \emph{2018 IEEE/CVF Conference on Computer Vision and Pattern
  Recognition}, Jun 2018{\natexlab{a}}.
\newblock \doi{10.1109/cvpr.2018.00813}.
\newblock URL \url{http://dx.doi.org/10.1109/CVPR.2018.00813}.

\bibitem[Wang and Hebert(2016)]{wang2016learning2learn}
Yu-Xiong Wang and Martial Hebert.
\newblock Learning to learn: Model regression networks for easy small sample
  learning.
\newblock In \emph{European Conference on Computer Vision}, pages 616--634.
  Springer, 2016.

\bibitem[Wang et~al.(2018{\natexlab{b}})Wang, Girshick, Hebert, and
  Hariharan]{Wang_2018}
Yu-Xiong Wang, Ross Girshick, Martial Hebert, and Bharath Hariharan.
\newblock Low-shot learning from imaginary data.
\newblock \emph{2018 IEEE/CVF Conference on Computer Vision and Pattern
  Recognition}, Jun 2018{\natexlab{b}}.
\newblock \doi{10.1109/cvpr.2018.00760}.
\newblock URL \url{http://dx.doi.org/10.1109/CVPR.2018.00760}.

\bibitem[Yang et~al.(2019)Yang, Meng, Li, Wu, Xu, and Chen]{local-transform}
Yuwei Yang, Fanman Meng, Hongliang Li, Qingbo Wu, Xiaolong Xu, and Shuai Chen.
\newblock A new local transformation module for few-shot segmentation.
\newblock \emph{Lecture Notes in Computer Science}, page 76–87, Dec 2019.
\newblock ISSN 1611-3349.
\newblock \doi{10.1007/978-3-030-37734-2\_7}.
\newblock URL \url{http://dx.doi.org/10.1007/978-3-030-37734-2\_7}.

\bibitem[Zhang et~al.(2019{\natexlab{a}})Zhang, Lin, Liu, Guo, Wu, and
  Yao]{pgnet}
Chenghui Zhang, Guosheng Lin, Fayao Liu, Jiushuang Guo, Qingyao Wu, and Rui
  Yao.
\newblock Pyramid graph networks with connection attentions for region-based
  one-shot semantic segmentation.
\newblock In \emph{ICCV 2019}, 2019{\natexlab{a}}.

\bibitem[Zhang et~al.(2019{\natexlab{b}})Zhang, Lin, Liu, Yao, and Shen]{canet}
Chi Zhang, Guosheng Lin, Fayao Liu, Rui Yao, and Chunhua Shen.
\newblock Canet: Class-agnostic segmentation networks with iterative refinement
  and attentive few-shot learning.
\newblock \emph{2019 IEEE/CVF Conference on Computer Vision and Pattern
  Recognition (CVPR)}, Jun 2019{\natexlab{b}}.
\newblock \doi{10.1109/cvpr.2019.00536}.
\newblock URL \url{http://dx.doi.org/10.1109/CVPR.2019.00536}.

\bibitem[Zhang et~al.(2018)Zhang, Wei, Yang, and Huang]{sgone}
Xiaolin Zhang, Yunchao Wei, Yi~Yang, and Thomas Huang.
\newblock Sg-one: Similarity guidance network for one-shot semantic
  segmentation, 2018.

\bibitem[Zhao et~al.(2017)Zhao, Shi, Qi, Wang, and Jia]{pspnet}
Hengshuang Zhao, Jianping Shi, Xiaojuan Qi, Xiaogang Wang, and Jiaya Jia.
\newblock Pyramid scene parsing network.
\newblock \emph{2017 IEEE Conference on Computer Vision and Pattern Recognition
  (CVPR)}, Jul 2017.
\newblock \doi{10.1109/cvpr.2017.660}.
\newblock URL \url{http://dx.doi.org/10.1109/CVPR.2017.660}.

\bibitem[Zhou et~al.(2020)Zhou, Zhou, Zhang, Yi, and Ouyang]{zhou2020cheaper}
Dongzhan Zhou, Xinchi Zhou, Hongwen Zhang, Shuai Yi, and Wanli Ouyang.
\newblock Cheaper pre-training lunch: An efficient paradigm for object
  detection.
\newblock \emph{arXiv preprint arXiv:2004.12178}, 2020.

\end{thebibliography}

\end{document}